\documentclass[letterpaper, 10 pt, conference]{ieeeconf}  

\IEEEoverridecommandlockouts                              

\usepackage{graphics} 
\usepackage{epsfig} 
\usepackage{times} 
\usepackage{amsmath} 
\usepackage{amssymb}  
\usepackage{cite}
\usepackage{booktabs}
\usepackage{graphicx}
\usepackage{diagbox}
\usepackage{float}
\usepackage[normalem]{ulem}
\usepackage{xcolor}

\title{\LARGE \bf
Reward Evolution with Graph-of-Thoughts: A Bi-Level Language Model Framework for Reinforcement Learning
}

\author{
Changwei Yao$^{1}$, Xinzi Liu$^{2}$, Chen Li$^{1}$, and Marios Savvides$^{1\dagger}$\\
$^{1}$Carnegie Mellon University, $^{2}$University of Tokyo
\thanks{$\dagger$Corresponding author: Marios Savvides.}
\thanks{$^{1}$Carnegie Mellon University, Pittsburgh, PA 15213, United States
        {\tt\small \{changwey, chenli4, marioss\}@andrew.cmu.edu}}%
\thanks{$^{2}$The University of Tokyo, 7-3-1 Hongo, Bunkyo-ku, Tokyo 113-8654, Japan
        {\tt\small liuxz@g.ecc.u-tokyo.ac.jp}}%
}

\begin{document}

\maketitle
\thispagestyle{empty}
\pagestyle{empty}

\begin{abstract}

Designing effective reward functions remains a major challenge in reinforcement learning (RL), often requiring considerable human expertise and iterative refinement. Recent advances leverage Large Language Models (LLMs) for automated reward design, but these approaches are limited by hallucinations, reliance on human feedback, and challenges with handling complex, multi-step tasks. In this work, we introduce Reward Evolution with Graph-of-Thoughts (RE-GoT), a novel bi-level framework that enhances LLMs with structured graph-based reasoning and integrates Visual Language Models (VLMs) for automated rollout evaluation. RE-GoT first decomposes tasks into text-attributed graphs, enabling comprehensive analysis and reward function generation, and then iteratively refines rewards using visual feedback from VLMs without human intervention. Extensive experiments on 10 RoboGen and 4 ManiSkill2 tasks demonstrate that RE-GoT consistently outperforms existing LLM-based baselines. On RoboGen, our method improves average task success rates by 32.25\%, with notable gains on complex multi-step tasks. On ManiSkill2, RE-GoT achieves an average success rate of 93.73\% across four diverse manipulation tasks, significantly surpassing prior LLM-based approaches and even exceeding expert-designed rewards. Our results indicate that combining LLMs and VLMs with graph-of-thoughts reasoning provides a scalable and effective solution for autonomous reward evolution in RL.

\end{abstract}

\section{INTRODUCTION}

Reinforcement learning (RL) has been successfully deployed in many fields, especially in robotics where agents learn complex skills in open environments~\cite{rudin2022learning, rajeswaran2017learning, kalashnikov2018scalable}. Simultaneously, emerging simulators enhance the simulation world with more details to eliminate the sim-to-real gap and expedite the RL training process. However, one of the key challenges of applying RL is designing an appropriate reward function that will lead to the desired behavior of robots and requiring extensive tuning to optimize the efficacy, called reward engineering. To mitigate this problem, inverse RL~\cite{wulfmeier2015maximum, ho2016generative} and preference-based RL~\cite{lee2021pebble,christiano2017deep} were designed in prior work. But it still requires great efforts in collecting demonstrations and suffers from limited reward diversity and too strong bias from experts, and thus is prone to overfitting and less capable of generalization.

Recently, Large Language Models (LLMs) have demonstrated remarkable capabilities in reasoning and coding, enabling them to analyze complex problems, generate efficient solutions, and even assist in debugging and optimization. Thereby, prior work has studied replacing human supervision by LLMs to write code-based reward functions~\cite{ma2023eureka,xie2023text2reward}. However, one of the major challenges with LLMs is their tendency to hallucinate, generating incorrect or misleading information with high confidence, particularly when solving complex problems. Since it occurs frequently~\cite{banerjee2025llms} and reduces the efficacy of reward functions, some studies mitigate this limitation with human feedback in the loop, which still relies on expert knowledge. Furthermore, directly generating reward functions from LLMs is often insufficient for long-horizon, multi-step tasks due to limited structured reasoning over the complex topology of robot tasks.

\begin{figure}[t]
    \centering
    \includegraphics[width=\linewidth]{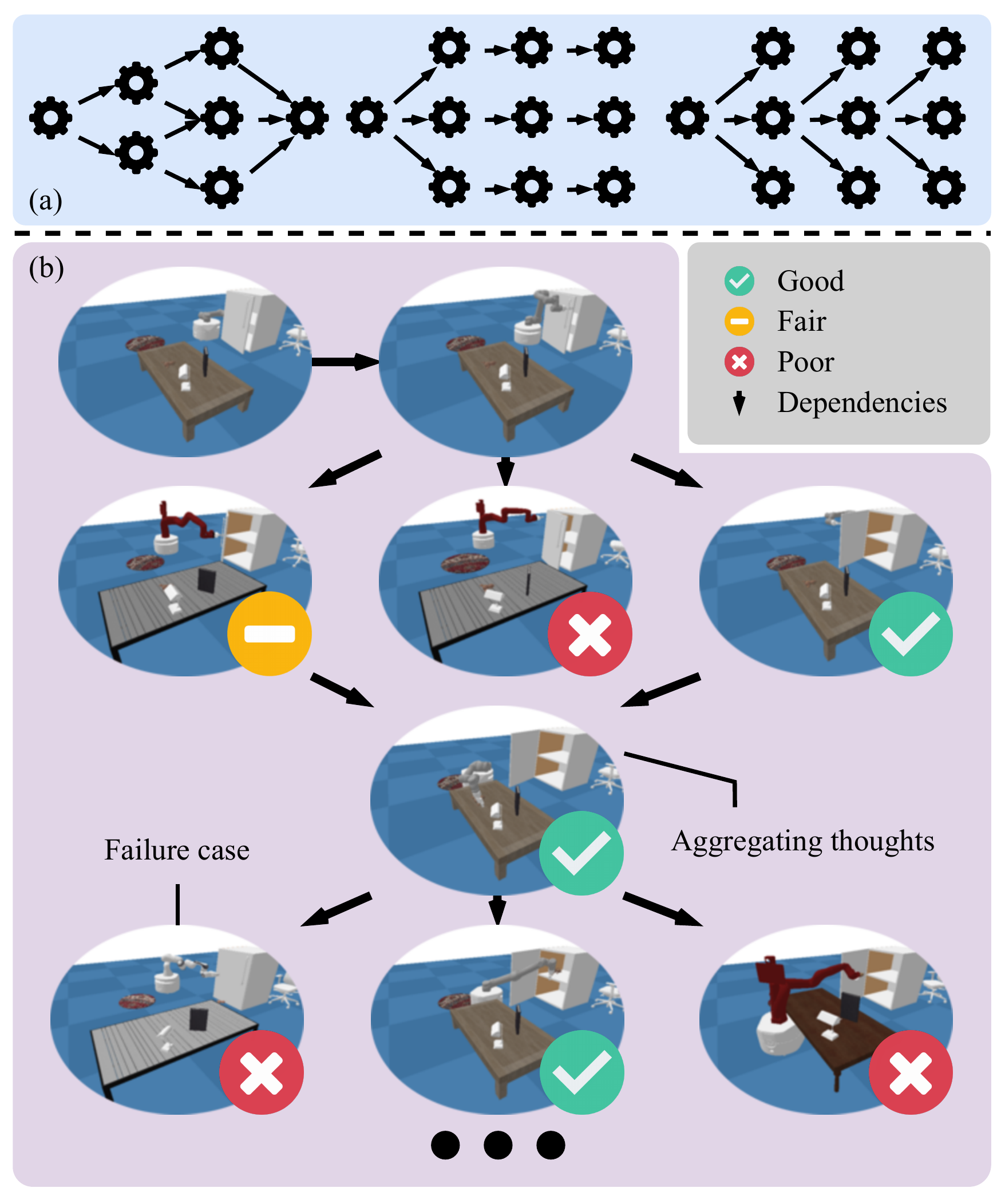}
    \caption{Conceptual illustration of GoT. (a) Three general GoT examples, where each node is a thought. (b) GoT applied to a manipulation task named \textit{Store Item in Storage}: nodes represent sub-goal stages, and edges represent robot behaviors for transitioning between stages, both with detailed textual descriptions.}
    \label{fig:intro}
\end{figure}

To further enhance LLMs' reasoning capabilities and reduce reliance on human supervision, recent advancements such as Chain-of-Thought (CoT)~\cite{wei2022chain,li2024chain,mitra2024compositional} and Graph-of-Thought (GoT)~\cite{besta2024graph,yao2024got} have been introduced. These methods enable structured reasoning, with GoT particularly excelling in handling complex decision-making tasks by organizing information in a graph-based manner rather than a linear sequence. By leveraging GoT as shown in Fig.~\ref{fig:intro}, LLMs can explore multiple reasoning paths simultaneously, making them more effective in generating well-structured and coherent reward functions. This structured approach not only mitigates hallucinations but also improves the adaptability of RL systems in dynamic environments.

In this work, we aim to eliminate human intervention in mitigating LLM hallucinations by proposing Reward Evolution with Graph-of-Thoughts (RE-GoT), a framework that enables LLMs to construct a graph-based representation of the task's solution before generating reward functions.
We posit that the performance on long-horizon tasks is not merely a function of model scale, but is significantly bolstered by an architectural scaffolding that imposes necessary structural constraints to effectively organize and ground this knowledge. RE-GoT addresses this by enforcing a bi-level reasoning topology.
The lower level only needs the environment codes and a text description of the task goal, mapping out a graph-based solution to the final goal first and designing the reward functions afterwards. Meanwhile, the upper level requires only a rollout video of the trained RL agent and leverages feedback from vision-language foundation models (VLMs) trained on large-scale multimodal datasets, thereby replacing human supervision, mitigating bias, and enabling a closed-loop update process. Furthermore, rather than instructing LLMs to generate the reward functions directly~\cite{ma2023eureka,xie2023text2reward,li2024auto}, equipping them with the GoT ability enables LLMs to consider more comprehensively, particularly when dealing with long-horizon, multi-step, and complex tasks with multiple solutions, making this difference significant.

In summary, our key contributions are as follows: a) we introduce the first to leverage GoT in automatic adaptive reward generation, enabling more structured and effective task decomposition, b) we propose a bi-level framework that improves reward function learning by leveraging VLMs feedback from video demonstrations without human, allowing for more adaptive policy training, and c) we demonstrate the adaptability of our approach across different platforms, showcasing its potential as a mobile tool through extensive comparative analysis and ablation studies.

\begin{figure*}[t]
    \centering
    \includegraphics[width=\textwidth]{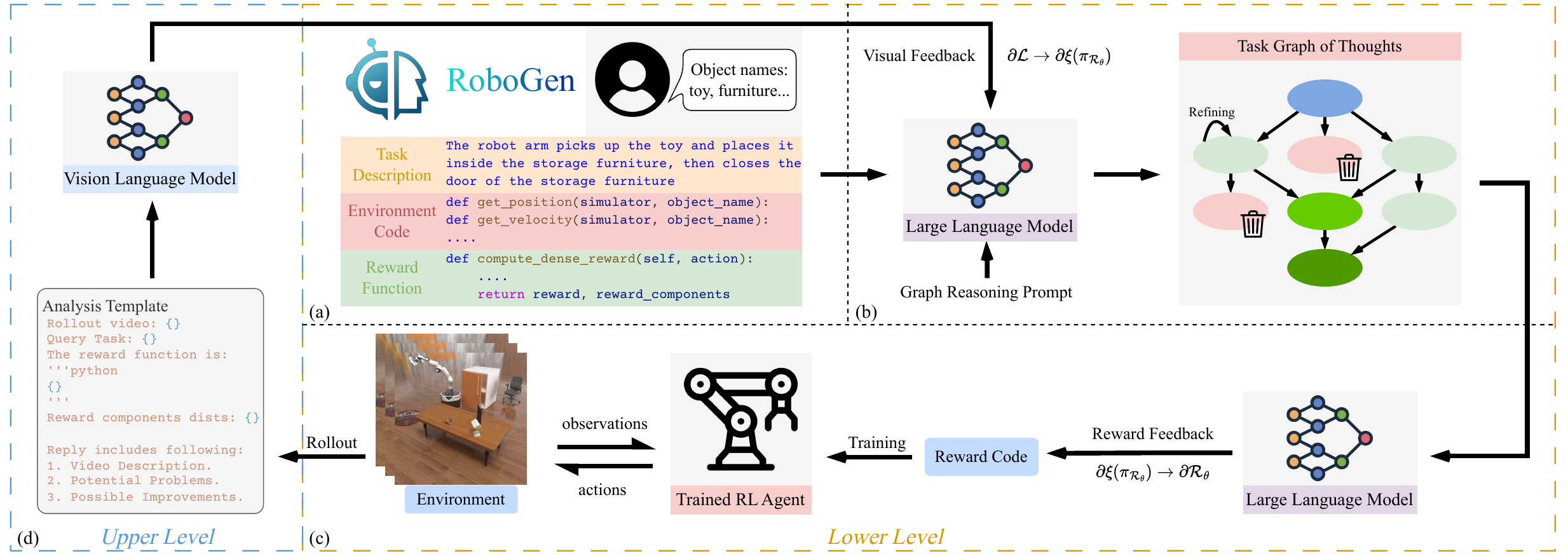}
    \caption{Overview of the RE-GoT framework. The upper-level evaluates rollout videos using VLMs to provide visual feedback, while the lower-level refines reward functions using LLMs with a graph-based reasoning approach. (a) It prompts LLMs with the environment abstraction to connect to the robotics system. (b) LLMs decompose the task into a text-attributed graph. (c) Given the graph structure and visual feedback, LLMs refine the reward function. (d) VLMs analyze the rollout videos to provide structured feedback on the trained RL agent.}
    \label{fig:pipeline}
\end{figure*}

\section{RELATED WORK}

\noindent \textbf{Reward Design.} Reward engineering remains a significant challenge in RL~\cite{gupta2022unpacking,laud2004theory,singh2009rewards}. The performance of an RL agent trained via reward shaping is heavily dependent on the quality of its reward function. Many studies investigate the construction of high-quality reward functions from diverse perspectives. Inverse Reinforcement Learning (IRL)~\cite{ng2000algorithms,ziebart2008maximum,fu2017learning} was introduced to infer rewards autonomously from expert demonstrations, but it still suffers from high data acquisition costs and produces black-box reward models that lack interpretability and are difficult to adapt. Preference-based learning~\cite{christiano2017deep,ibarz2018reward}, which builds reward functions by having humans compare different behaviors or trajectories, also fails to eliminate the reliance on costly annotations and introduces noise due to inconsistent individual preferences. In contrast, our RE-GoT framework requires only a few examples to adapt across tasks, effectively overcoming the high data and annotation costs of prior methods.

\noindent \textbf{LLMs and Prompting for RL.} LLMs have recently demonstrated significant potential in RL. Early works~\cite{du2023vision,du2023guiding,fan2022minedojo,karamcheti2023language,kwon2023reward} leverage pretrained foundation models to generate reward signals for RL agents, but these approaches often require frequent queries to LLMs, resulting in high token consumption and reduced training efficiency due to the large number of environment samples needed. More recent studies~\cite{yu2023language,xie2023text2reward,ma2023eureka,li2024auto,sun2024large} focus on enabling robots to acquire low-level skills by designing effective reward functions, while others~\cite{zeng2024learning,wang2023robogen} introduce CoT prompting to enhance LLM reasoning for improved reward design. However, these methods still face challenges in adapting to complex, multi-step tasks. Instead, RE-GoT utilizes graph-based thinking, similar to how humans solve complex problems, empowering LLMs with comprehensive analysis and reasoning capabilities to generate more effective rewards. Furthermore, RE-GoT replaces human intervention with VLMs, accelerating the reward search process.

\section{PRELIMINARY}

\textbf{Problem Setup.} We define our robot agent as a Markov Decision Process (MDP), represented by the tuple $\mathcal{M}=\left(\mathcal{S},\mathcal{A},\mathcal{R},\mathcal{P},\gamma, \rho_0\right)$, where the environment consists of a state space $s\in\mathcal{S}$ and an action space $a\in\mathcal{A}$. The transition model $\mathcal{P}(s'\mid s,a)$ governs the environment dynamics while $\rho_0$, $\mathcal{R}$ and $\gamma$ are the distribution of the initial robot state, the environment's reward function, and the discount factor, respectively. We aim to get an optimal policy $a_t\in\pi_{\mathcal{R}_\theta}\left(a_t\mid s_t\right)$ given the parameterized reward function $\mathcal{R}_\theta:\mathcal{S}\times\mathcal{A}\rightarrow\mathbb{R}$ by maximizing the expected reward:
\begin{equation}
    \pi_{\mathcal{R}_\theta}=\displaystyle{\arg\max_{\pi}}\ \mathbb{E}_{\pi,\mathcal{M}}\left[\sum_{t=0}^T \gamma^t{\mathcal{R}_\theta}\left(s_t,a_t\right)\right]
\end{equation}
where $T$ denotes the trajectory length, and we define the rollout trajectory over the optimal policy $\pi_{\mathcal{R}_\theta}$ as follows:
\begin{equation}
    \xi\left(\pi_{\mathcal{R}_\theta}\right)\sim\rho_0\left(s_0\right)\prod_{t=0}^{T-1}\mathcal{P}\left(s_{t+1}|s_t,a_t\right)\pi\left(a_t|s_t\right)
\end{equation}

We consider a framework where LLMs are provided with a textual description $\mathcal{T}$ and a graph structure $\mathcal{G}$ of the task, while VLMs receive the rollout video of a trained RL agent $\mathcal{D}\left(\xi\left(\pi_{\mathcal{R}_\theta}\right)\right),\;\xi\left(\pi_{\mathcal{R}_\theta}\right)\in\Xi$. Given an observation model $\mathcal{D}:\Xi\rightarrow\mathcal{D}$ mapping rollout trajectory to the observation space, our objective is to minimize the difference between the expected and actual behaviors:
\begin{equation}
    \displaystyle{\min_{\mathcal{R}_\theta}}\ \underset{\xi(\pi_{\mathcal{R}_\theta})}{\mathbb{E}}\mathcal{L}\left[E_{LLM}(\mathcal{T},\mathcal{G}),E_{VLM}(\mathcal{D}(\xi(\pi_{\mathcal{R}_\theta})))\right]
    \label{obj_func}
\end{equation}

\textbf{Classic Gradient-Based Optimization.} The optimization in loss function~\eqref{obj_func} represents a bi-level problem~\cite{mahesheka2024language}, where the upper-level optimization minimizes the visual loss of $\xi(\pi_{\mathcal{R}_\theta})$ under policy $\pi_{\mathcal{R}_\theta}$, and the lower-level solves the RL problem to obtain $\pi_{\mathcal{R}_\theta}$. Based on classical gradient-based method, we optimize the object by computing the reward gradient:
\begin{equation}
    \nabla_{\mathcal{R}_\theta} \mathcal{L} = \frac{\partial \mathcal{L}}{\partial \xi(\pi_{\mathcal{R}_\theta})} \cdot \frac{\partial \xi(\pi_{\mathcal{R}_\theta})}{\partial {\mathcal{R}_\theta}}
    \label{eq:gradient}
\end{equation}

However, before computing the gradient, we need to make sure the loss $\mathcal{L}$ and the observation model $\mathcal{D}$ be explicitly defined and differentiable, which is not feasible due to the need for data preprocessing and explicit modeling. To address this, we propose leveraging the capabilities of VLMs and LLMs as a gradient-free alternative to solving~\eqref{eq:gradient}. Instead of differentiating through RL training, we frame the problem as aligning expert task knowledge, provided by LLMs in the form of text $\mathcal{T}$ and graph structure $\mathcal{G}$, with the visual rollout $\mathcal{D}(\xi(\pi_{\mathcal{R}_\theta}))$ evaluated by VLMs. This avoids the need for explicitly modeling $\mathcal{L}$ and $\mathcal{D}$, allowing reward learning to be guided by high-level expert priors rather than relying on direct gradient propagation through the reinforcement learning process.

\section{METHOD}

\subsection{RE-GoT Overview}
Based on the reward gradient~\eqref{eq:gradient}, for learning a reward with a better performance, we search the space in the direction with the two stages via chain rule, where 1) Visual feedback $\partial \mathcal{L} \rightarrow \partial \xi(\pi_{\mathcal{R}_\theta})$, from structured embedding loss to robot behaviors. 2) Reward update $\partial \xi(\pi_{\mathcal{R}_\theta}) \rightarrow \partial {\mathcal{R}_\theta}$, from robot behaviors to reward functions. Specifically, the first stage is to minimize the behavior loss by visual feedback, while the second stage informs how to update reward in response to the robot behavior improvement. As it illustrates the RE-GoT framework in Fig.~\ref{fig:pipeline}, following the bi-level optimization, we divide our system into the upper-level named rollout evaluation (Section~\ref{sec:upper}), which focuses on evaluating the rollout videos and providing visual feedback, and the lower-level named reward refinement (Section~\ref{sec:lower}), which aims to refine the reward function based on the graph structure of the task and the feedback from the upper-level.

\subsection{Upper-Level: Rollout Evaluation.} 
\label{sec:upper}
The upper level leverages VLMs to evaluate the performance of RL agents by analyzing rollout videos given the analysis template shown in Fig.~\ref{fig:pipeline} (d). After the RL agent is trained with the current reward function, several rollout videos are generated to capture its behavior. The VLMs act as an automated evaluator, processing these videos to provide structured, textual feedback on task completion, failure modes, and areas for improvement. This feedback is used to identify discrepancies between the agent's behavior and the intended task objectives, guiding subsequent reward refinement. By automating the evaluation process and reducing reliance on human feedback, this stage accelerates the reward evolution loop and ensures that the agent's learning is aligned with high-level task goals.

\subsection{Lower-Level: Reward Refinement.} 
\label{sec:lower}
\subsubsection{System initialization}
Component Fig.~\ref{fig:pipeline} (a) enables the connection of LLMs to the robotics simulator via the environment abstraction, which will be executed only once from the beginning. This abstraction encompasses the observation and action spaces, a detailed task description including all relevant objects, initial reward functions, and a set of callable Python APIs. These APIs provide access to essential environment information, such as the robot's position, object poses, collision detection, etc.
Following the principles of task decomposition in RoboGen~\cite{wang2023robogen}, RE-GoT employs the LLM to categorize each substep as either \textit{primitive} or \textit{reward} according to its control complexity and analytical solvability. Primitive substeps are defined as motions with low contact complexity, which can be reliably executed via motion planning with simulator APIs. In contrast, reward substeps are designated for stages involving high-dimensional physical interactions where analytical modeling is infeasible and reinforcement learning is required to acquire adaptive policies.

\subsubsection{Text-attributed graph construction} 
\begin{figure}[H]
    \centering
    \includegraphics[width=\linewidth]{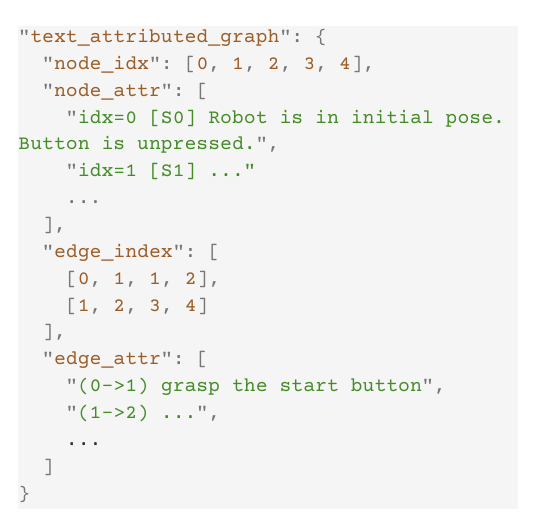}
    \caption{Example of the text-attributed graph for Press the Start Button, where $S_i$ indicates the index of the sub-goal.}
    \label{fig:text-att}
\end{figure}

To better leverage the reasoning capabilities of LLMs, Fig.~\ref{fig:pipeline} (b) first enables LLMs to decompose the task into a structured graph representation, providing an organized and interpretable view of the solution space. However, LLMs may sometimes generate incorrect or misleading outputs. To address this, we employ heuristic rules and in-context learning with a few examples to guide the LLM during the decomposition process. Once provided with the task description, the LLM constructs a text-attributed graph, which is then used for subsequent reward refinement. An example of the text-attributed graph represented as its GoT is shown in Fig.~\ref{fig:text-att}.

Formally, we define a text-attributed graph $\mathcal{G} = (\mathcal{V}, \mathcal{E}, \mathcal{T}_v, \mathcal{T}_e)$, where $\mathcal{V}$ is the set of nodes representing stages, $\mathcal{E}$ is the set of edges denoting transitions between stages, and $\mathcal{T}_v$, $\mathcal{T}_e$ are the sets of attributes associated with each node and edge, respectively. Although the number of steps to achieve the task is fixed for each task, each sub-goal may yield multiple possible outcomes due to variations in actions and environments. The node attributes $\mathcal{T}_v$ describe each stage using the robot, objects, and environment status, while the edge attributes $\mathcal{T}_e$ characterize the robot behaviors required for stage transitions. Similar to human problem-solving, there are often multiple ways to reach the next sub-goal, with varying efficiency and likelihood of success. Unlike prior work~\cite{besta2024graph}, our approach does not require manual design of the strategy graph or repeated LLM queries for each node. Instead, we leverage the LLM once to generate the entire graph of thoughts for the task, resulting in a more efficient and resource-saving process.

\subsubsection{Reward function refinement} Rather than directly generating the reward function, Fig.~\ref{fig:pipeline} (c) uses the interfaces from the robotics system as the prior knowledge mentioned in component (a). With all the information fed into the LLM, the reward function refinement process becomes more targeted, allowing the LLM to better understand the system's capabilities and constraints. This enables the LLM to generate reward functions that are both practical and aligned with the specific task objectives, ensuring the robot's behavior is effectively guided by the reward signals.

As we get the visual feedback from VLMs and the text-attributed graph of thoughts from LLMs, we prompt them both with current used reward function to LLMs to refine the reward function in two aspects: 
1) Add more constraints, remove redundant constraints or modify current function form to make it more suitable for the current task. 2) For each component within the reward function, search the best weights $\theta$ for each component to make the reward function $\mathcal{R}_\theta$ more effective.

\subsection{Implementation Details}

In our framework, we utilize gpt-4o-2024-08-06 as our core reasoning engine, prompted as an expert in robotics and reinforcement learning. The LLM performs a single-query task decomposition into a text-attributed graph consisting of vertices and edges. For each reward substep, the LLM generates reward code, selects the optimal action space, and defines verifiable success conditions based on simulator APIs. For visual supervision, gemini-1.5-pro serves as the VLM evaluator, receiving the video alongside the current reward function and a statistical distribution (maximum, mean, minimum, and standard deviation) of individual reward components. The VLM then provides structured semantic feedback, including a video description, identification of motion problems, and specific reward redesign suggestions.

For policy learning, we adopt Soft Actor-Critic (SAC) as the primary reinforcement learning algorithm for most tasks, while Proximal Policy Optimization (PPO) is employed specifically for the \textit{PickCube} task in ManiSkill2 to ensure stable convergence. Following the protocol in RoboGen, we conduct training for $1\times10^6$ environment steps per reward substep to ensure policy proficiency. The observation space is composed of low-level environment states, including 6D object poses and joint angles of articulated parts, while the action space is defined as 6D end-effector control. All experiments were conducted on one NVIDIA RTX 4070 Ti Super GPU.

\begin{figure*}[t]
    \centering
    \includegraphics[width=\textwidth]{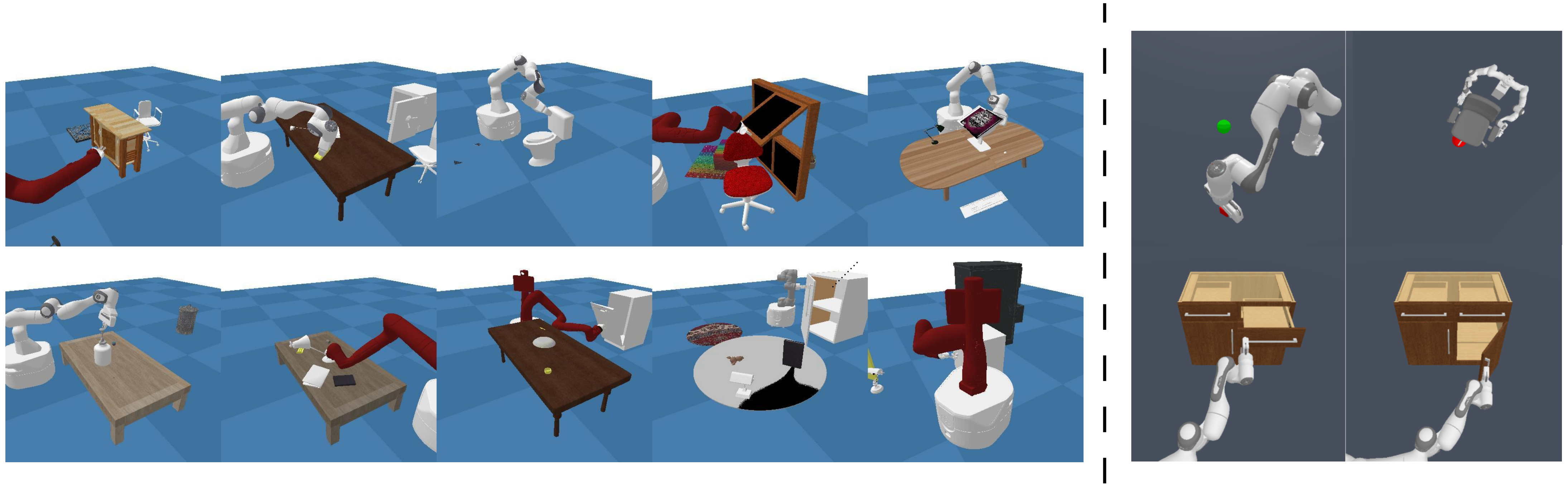}
    \caption{Evaluation environments. Ten tasks from RoboGen on the left: \textit{Open both table doors}, \textit{Retrieve item from safe}, \textit{Flush toilet}, \textit{Close window}, \textit{Tilt display screen}, \textit{Close dispenser lid}, \textit{Turn on Lamp}, \textit{Load dish into dishwasher}, \textit{Store item into storage}, and \textit{Rotate safe knob}. Four tasks from ManiSkill2 on the right: \textit{PickCube}, \textit{OpenCabinetDrawer}, \textit{OpenCabinetDoor}, and \textit{PushChair}.}
    \label{fig:task}
\end{figure*}

\section{EXPERIMENTS}

\subsection{Evaluation Setup}

\subsubsection{Objectives} To evaluate our proposed framework, we investigate the following hypotheses through a series of experiments: 

$\mathcal{H}_1$ - Can this framework enhance reward functions to achieve effective agents across general manipulation tasks? 

$\mathcal{H}_2$ - How does this pipeline perform compared to other LLM-based unsupervised methods? 

$\mathcal{H}_3$ - How much impact does RE-GoT have on LLM reasoning compared to unstructured direct prompting, and does in-context learning really help improve the GoT ability?

$\mathcal{H}_4$ - Does in-context learning really help improve the GoT ability to achieve a more stable and precise reward function?

$\mathcal{H}_5$ - Can iterative updates on the graph of the task reduce the numerical imprecision and instability, thereby improving the efficacy? 

\subsubsection{Environment} We evaluate our framework on a diverse suite of 14 robotic manipulation tasks designed to simulate household interaction scenarios. These tasks are sourced from two prominent benchmarks: RoboGen~\cite{wang2023robogen} and Maniskill2~\cite{gu2023maniskill2}. From RoboGen, we adopt 10 tasks spanning a wide range of object interactions and actuator types. These tasks are implemented in PyBullet with extensive randomization across robots and objects properties. We additionally include 4 tasks from ManiSkill2 to further validate the generalization of our framework.

\subsubsection{Metrics} To evaluate skill learning performance, we use the success rate of the learned policy as the primary metric. This metric quantifies the proportion of successful attempts in achieving the desired task outcome, providing a clear indication of the effectiveness of the learned policy. We also report the mean episode length for each task to provide a more comprehensive understanding of the performance across different tasks. 

\subsection{Baseline}

We compare our method with three baseline approaches of reward generation with LLM:
a) \textbf{RoboGen}~\cite{wang2023robogen} integrates task proposal, scene generation, training supervision generation, and skill learning into one pipeline, obtaining a trained agent given task scheme or even only objects. It also provides numerous example tasks and allows users to generate custom ones.
b) \textbf{RewardSelfAlign}~\cite{zeng2024learning} (SA) iteratively refines LLM-generated reward functions through self-alignment, updating reward parameters by minimizing ranking discrepancies between LLM-preferred trajectories and the learned reward function.
c) \textbf{Text2Reward}~\cite{xie2023text2reward} (T2R) generates dense reward function code using zero-shot or few-shot method by providing the environment abstraction and task description to LLM and refines reward functions given the human feedback to LLM.

We compare the performance in various tasks in RoboGen to examine if RE-GoT could be generalized across different tasks for $\mathcal{H}_1$. In order to answer $\mathcal{H}_2$, we use the reward functions generated by T2R and SA, and reward functions generated by our system to train a RL agent respectively, comparing the training process. 

\begin{figure*}[t]
    \centering
    \includegraphics[width=\textwidth]{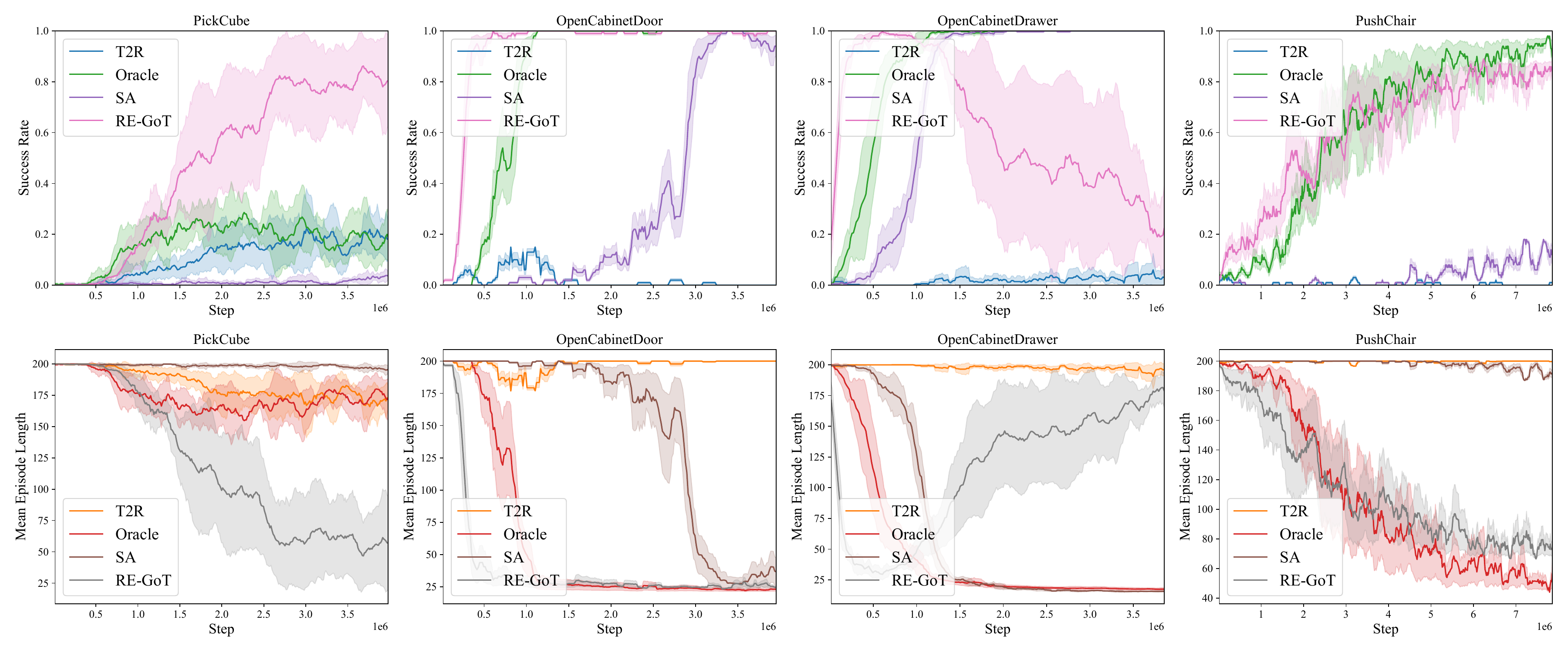}
    \caption{Success Rate \& Average Episode Length vs Exploration Steps on four ManiSkill2 tasks. The solid lines represent the mean, while the shaded areas indicate the standard error of the mean. Oracle means the expert-written reward function provided by the environment.}
    \label{fig:maniskill_success_rate}
\end{figure*}

\begin{table}[b]
    \centering
    \caption{SR (\%) comparison on RoboGen tasks}
    \label{tab:robogen_success_rate}
    \resizebox{\linewidth}{!}{%
        \begin{tabular}{c|lcc}
            \toprule
            \textbf{Substeps} & \textbf{Task} & \textbf{RoboGen} & \textbf{RE-GoT} \\
            \midrule
            2 & Close Dispenser Lid & 25.0$\pm$2.9 & \textbf{90.0$\pm$4.1} \\ 
              & Turn On Lamp & 25.0$\pm$6.5 & \textbf{80.0$\pm$0.0} \\ 
              & Rotate Safe Knob & 30.0$\pm$5.8 & \textbf{82.5$\pm$2.5} \\ 
            \midrule
            3 & Flush Toilet & 90.0$\pm$0.0 & \textbf{100.0$\pm$0.0} \\ 
              & Tilt Display Screen & 30.0$\pm$0.0 & \textbf{65.0$\pm$6.5} \\ 
            \midrule
            4 & Close Window & 80.0$\pm$0.0 & \textbf{100.0$\pm$0.0} \\ 
              & Open Both Table Doors & 80.0$\pm$0.0 & \textbf{100.0$\pm$0.0} \\ 
            \midrule
            6 & Load Dish into Dishwasher & \textbf{10.0$\pm$5.8} & \textbf{10.0$\pm$5.8} \\ 
              & Store Item in Storage & 10.0$\pm$4.1 & \textbf{47.5$\pm$2.5} \\ 
            \midrule
            8 & Retrieve Item from Safe & 42.5$\pm$2.5 & \textbf{70.0$\pm$0.0} \\ 
            \bottomrule
        \end{tabular}
    }
\end{table}

\subsection{Results and Analysis}
For evaluation, we report success rates across 10 tasks from RoboGen over 4 random seeds with 10 trials per seed (40 trials per task in total), comparing the performance of the evolved reward functions generated by RE-GoT against the original baselines provided by RoboGen.
This comparison highlights the impact of our framework in enhancing reward design for general robotic manipulation tasks, which is shown in Table~\ref{tab:robogen_success_rate}. We also evaluate four reward functions generated by T2R in zero-shot settings, as well as the best hyperparameter combinations produced by SA. Their performances are compared with ours in terms of success rate and average episode length, as shown in Fig.~\ref{fig:maniskill_success_rate}.

\subsubsection{The enhanced reward achieves an agent with better performance}
As shown in Table~\ref{tab:robogen_success_rate}, we classify the 10 tasks into 5 categories based on the number of substeps required to complete the task. The success rates of the reward functions generated directly by LLMs are generally lower than those from ours, i.e. an improvement of 32.25\% on average which supports $\mathcal{H}_1$. For the task Load Dish into Dishwasher, the randomization for the positions and orientations of the desk and the dishwasher always makes it impossible to open the dishwasher door completely by the robot, which explains no improvement on this task. We observe that while our method achieves the largest improvement on tasks with 2 substeps, its effectiveness remains consistent as the number of substeps increases from 3 to 8. This indicates that our method scales well with task complexity and maintains its advantage over the original reward functions, even in more demanding settings.

\subsubsection{RE-GoT performs better across different platforms against other LLM-based baselines}
To further evaluate the stability and generalization of RE-GoT and address $\mathcal{H}_2$, we conduct experiments on four environments from ManiSkill2, each with five different random seeds, as shown in Fig.~\ref{fig:maniskill_success_rate}. We observe that RE-GoT consistently outperforms the other two baselines across all tasks, achieving the best success rates of 86.20\%, 100.00\%, 99.67\%, and 89.00\%, respectively. While our method achieves performance comparable to the oracle, it notably improves the success rate for PickCube from 28.60\% to 86.20\%. However, we also find that the performance of our method on \textit{OpenCabinetDrawer} declines as training continues after step $1\times10^6$, we suspect that this may be because the VLM fails to analyze the rollout videos correctly and provide accurate feedback.

\subsection{Ablation Study}

\subsubsection{Effectiveness of the GoT Architecture}
\begin{figure}[H]
    \centering
    \includegraphics[width=\linewidth]{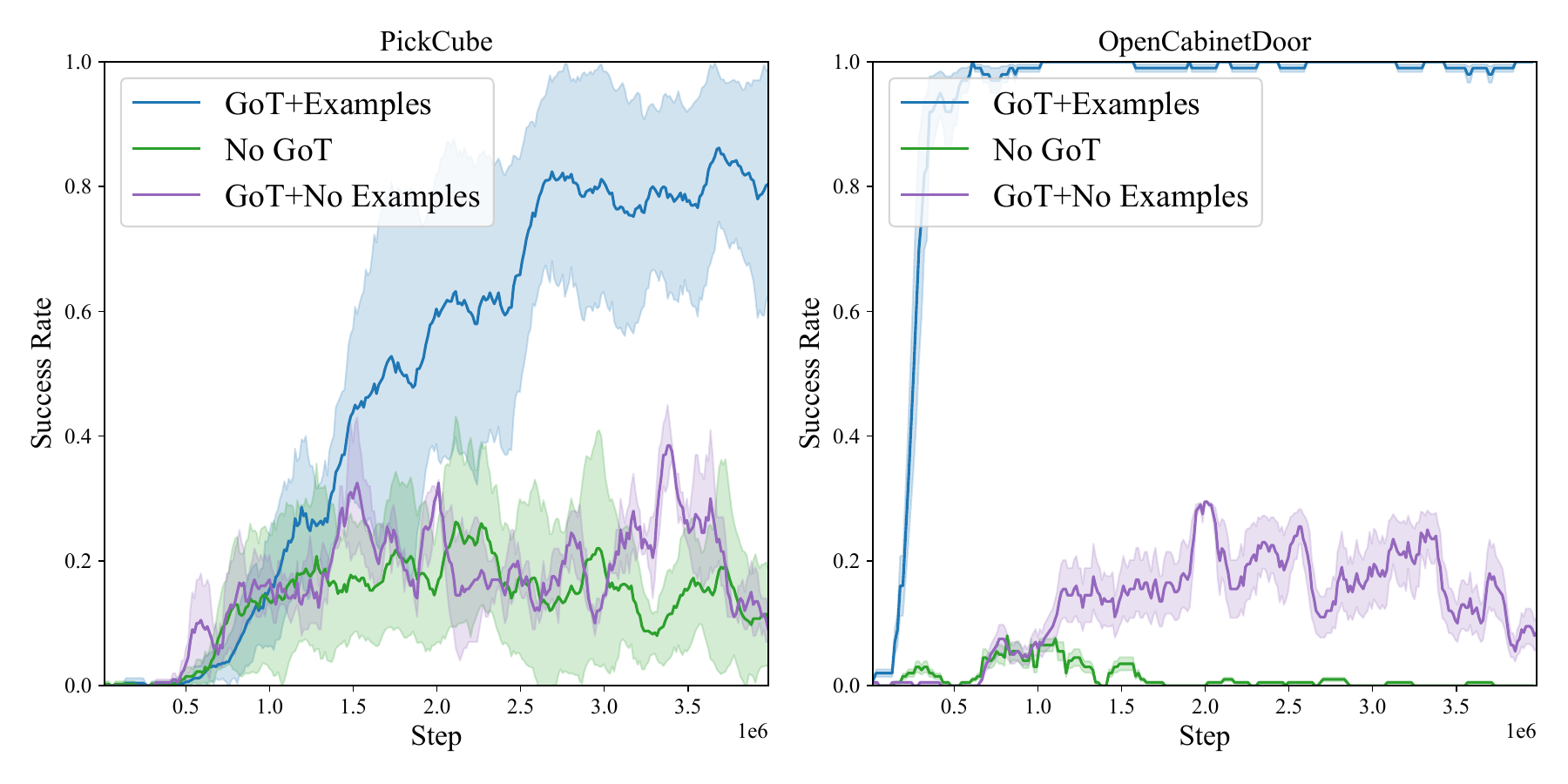}
    \caption{Effect of different prompting strategies on performance. LLMs achieve the best reward design with GoT and in-context examples.}
    \label{fig:ablation}
\end{figure}

To answer $\mathcal{H}_3$ and isolate the impact of our framework from the underlying model strength, we compare RE-GoT with a baseline without GoT. In this setting, the LLM is provided with identical environment abstractions and task descriptions but is restricted to unstructured direct prompting. As illustrated in Fig.~\ref{fig:ablation}, the performance on \textit{PickCube} and \textit{OpenCabinetDoor} reveals that there is nearly no improvement in success rates without the GoT structure. While these two representative cases are presented for visual clarity, we observed a consistent performance trend across the entire task suite. This result confirms that while state-of-the-art models possess significant internal knowledge, the implementation of a structured architectural scaffolding is a prerequisite for effectively organizing that knowledge to solve manipulation tasks.

\subsubsection{Effectiveness of in-context learning}
To evaluate the effectiveness of in-context learning, we compare the performance of LLMs in zero-shot and few-shot settings. In the few-shot setting, LLMs are provided with a few examples of task descriptions and graph structures as context, while in the zero-shot setting, no such examples are given. This comparison allows us to assess whether few-shot in-context learning improves the GoT ability, addressing $\mathcal{H}_4$. From Fig.~\ref{fig:ablation}, we can conclude that few-shot enables us to greatly improve the reasoning ability of LLMs to generate better rewards.

\subsubsection{Impact of continuous updating loop}
To determine whether periodic updates on the graph help create more stable and precise reward functions, we demonstrate the change in success rate as the number of iterations increases in the pipeline loop, which tries to figure out $\mathcal{H}_5$. As reported in Table~\ref{tab:iter}, we observe that the success rate of the tasks generally increases as the number of iterations increases and achieves its peak at 8 iterations. This indicates that the continuous updating loop helps to refine the reward functions and improve the performance of the RL agent.

\begin{table}[H]
    \centering
    \caption{SR (\%) of tasks as RE-GoT iterations increase}
    \label{tab:iter}
    \resizebox{\linewidth}{!}{
        \begin{tabular}{cccccc}
            \toprule
            \diagbox{\textbf{Task}}{\textbf{Iteration}} & \textbf{n=2} & \textbf{n=4} & \textbf{n=6} & \textbf{n=8}\\
            \midrule
            Turn On Lamp & 57.5$\pm$2.5 & 70.0$\pm$4.0 & 75.0$\pm$2.5 & 80$\pm$0.0\\
            Flush Toilet & 90.0$\pm$0.0 & 95.0$\pm$2.9 & 100.0$\pm$0.0 & 100.0$\pm$0.0\\
            Close Window & 90.0$\pm$0.0 & 92.5$\pm$2.5 & 100.0$\pm$0.0 & 100.0$\pm$0.0\\
            \bottomrule
        \end{tabular}
    }
\end{table}


\subsection{Efficiency Discussion}
Our method shifts the burden from non-scalable human reward engineering to scalable computational search in simulation. By decomposing long-horizon tasks into structured substeps via the GoT architecture, we significantly narrow the exploration space, enabling policy convergence within $1\times10^6$ steps per substep. Moreover, to manage costs and latency, we adopt a keyframe sampling strategy on the input video rollouts that minimizes token usage and ensures VLM feedback returns in under 10 seconds. Crucially, this reasoning process occurs strictly offline. Once the reward functions are evolved, the final policy can be deployed online without any LLM querying.

\subsection{Path to Physical Deployment Discussion}
Although current evaluations are conducted in simulation, the modular architecture of RE-GoT provides a viable path to physical deployment by bridging the sim-to-real gap. Following the pipeline established in Real2Sim-Eval~\cite{zhang2025real}, a high-fidelity digital twin can be reconstructed from real-world environments to ensure visual and physical consistency. Within this digital twin, SAM3D~\cite{chen2025sam} can be utilized for object-level scene reconstruction and segmentation, effectively replacing privileged simulator states with vision-based perception. Crucially, as the VLM-based evaluator operates directly on raw video streams rather than internal simulator code, it remains inherently domain-agnostic and capable of providing feedback in reconstructed real environments. This enables RE-GoT to function as an offline reward evolver within a digital twin, generating robust rewards that facilitate zero-shot or few-shot transfer to physical deployment.

\section{CONCLUSIONS \& LIMITATIONS}
In this work, we introduced RE-GoT, a novel bi-level framework that leverages GoT and LLMs for reward function evolution in reinforcement learning. By integrating structured graph-based reasoning and visual feedback from VLMs, our approach enables more effective task decomposition, adaptive reward refinement, and reduces reliance on human supervision. Extensive experiments on RoboGen and ManiSkill2 benchmarks demonstrate that RE-GoT consistently outperforms existing LLM-based baselines, achieving higher success rates and better generalization across diverse robotic manipulation tasks. Ablation studies further validate the importance of GoT prompting, in-context learning, and continuous reward refinement. Our results highlight the potential of combining LLMs and VLMs with graph-based reasoning to advance autonomous reward design and improve the scalability and adaptability of RL systems in complex environments.

RE-GoT is not without limitations. The performance of the framework is highly dependent on the quality of the LLMs and VLMs used, as well as the accuracy of the task description and graph structure provided. In cases where the LLMs generate incorrect or misleading outputs, the effectiveness of the reward functions may be compromised. Additionally, while our approach reduces reliance on human supervision, it still requires some level of expert knowledge to define the task description and graph structure. Future work could explore the integration of more advanced LLMs and VLMs for deployment on real robots, as well as the development of more robust prompting strategies to further enhance the performance and generalization of RE-GoT across a wider range of tasks and environments. Furthermore, we plan to create an extensive GoT dataset of different tasks and fine-tune the LLMs to improve the performance of our framework.







\bibliographystyle{IEEEtran}
\bibliography{reference}

\end{document}